# Fine-Tashkeel: Finetuning Byte-Level Models for Accurate Arabic Text Diacritization

**Bashar Al-Rfooh**
bashar@alrfou.com

**Gheith Abandah**
abandah@ju.edu.jo

**Rami Al-Rfou**
rmyeid@google.com

## Abstract

Most of previous work on learning diacritization of the Arabic language relied on training models from scratch. In this paper, we investigate how to leverage pre-trained language models to learn diacritization. We finetune token-free pre-trained multilingual models (ByT5) to learn to predict and insert missing diacritics in Arabic text, a complex task that requires understanding the sentence semantics and the morphological structure of the tokens. We show that we can achieve state-of-the-art on the diacritization task with minimal amount of training and no feature engineering, reducing `WER` by 40%. We release our finetuned models for the greater benefit of the researchers in the community.

## 1 Introduction

Arabic has an Abjad writing system where only the consonants and long vowels are being written (de Voogt and Quack, 2012). Later modifications to the writing system introduced short vowels in the form of diacritics. These diacritics are essential to disambiguate the meaning of the text. For example, many Arabic words are homographs where multiple words have the same spelling and diacritics can disambiguate them. However, many native speakers do not include these diacritics, saving time and effort, in their day-to-day writing assuming that the correct meaning can be inferred from the context. Automatic diacritization is of great benefit as a form of suggested grammar correction which the user can accept. The diacritized text will be easier to read especially for non-native speakers (See Table 1). Moreover, text with diacritization is easier to process by text-to-speech (TTS) systems.

| **Arabic** | **Translation** |
|---|---|
| أكلتُ السمكة حتى رأسها. | – |
| أكلتُ السمكة حتى رأسَها. | I have eaten the whole fish including its head. |
| أكلتُ السمكة حتى رأسُها. | I have eaten the whole fish except the head. |

Table 1: Effect of diacritics on sentence meaning.

Previous efforts on learning a statistical model for automatic diacritization relied on training machine learning models initialized randomly (Zitouni and Sarikaya, 2009). Pre-trained models such as BERT (Devlin et al., 2018), T5 (Raffel et al., 2020), GPT (Radford et al., 2018) has received wide adoption from the NLP community (Mikolov et al., 2013; Peters et al., 2017). However, those models, while easy to finetune for a wide range of downstream tasks, have been pretrained mainly on English corpora, limiting their ability to be used for other languages. mT5 expands the corpora of the pretraining stage of T5 models from English to 100+ languages (Xue et al., 2021b). Given the closed vocabulary approach used to segment the multilingual text using sentence-piece (Kudo and Richardson, 2018), capacity assigned to each language is variable. Another approach is to pretrain monolingual models such as AraT5 (Nagoudi et al., 2022). ByT5 simplifies mT5 models by replacing the closed vocabulary approach with an open one, where the network itself is responsible for learning the appropriate segments of a language from its utf-8 byte input sequences.

We adopt ByT5 models as our foundational model since our predictions are based on character level. We model our problem as a sequence-to-sequence generation problem where the input is a sequence of Arabic characters without diacritics presented to the network after being encoded into utf-8. The target output is the same sequence of characters interleaved with the predicted diacritics. With a small number of finetuning steps (≤15K) we are able to leverage the multilingual capabilities of the network to learn automatic Arabic diacritization achieving state-of-the-art results. To summarize our contributions:

- We leverage large pretrained models to learn the task of diacritization achieving state-of-the-art results
- We study the effect of data quality and size on the fientuning process, devising a curriculum that utilizes quality and size of training data.
- We study the impact of scaling our pretrained models on the performance of diacritization.

## 2 Related Work

In this Section, we will discuss the previous work on machine learning based Arabic diacritization, large language models, and character level modeling.

**Arabic Diacritization** have been thoroughly studied through full-supervised approaches. Recent works considered the problem of diacritizing Arabic text as a classification problem like (Karim and Abandah, 2021; Madhfar and Qamar, 2020; Barqawi, 2017).Another approach is to model the problem as a translation using a sequence to sequence model as have been proposed by (Mubarak et al., 2019). In both approaches, the network is initialized randomly and does not leverage any unsupervised training nor utilize any large corpora. (Stankevičius et al., 2022) utilized the pre-trained ByT5 model to recover diacritical marks in 13 Latin-script languages and achieved competitive results, within 1% of the current state-of-the-art results. On the other hand, our research accomplished a state-of-the-art result in diacritizing Arabic, leading to a minimum 40% reduction in Word Error Rate (WER).

**Character level modeling** adopts a token-free approach where the vocabulary is not a closed finite set of segments and words. This eliminates the out-of-vocabulary (OOV) problem which tends to be severe in languages with complex morphology such as Arabic. Choe et al. (2019) shows that character level language models can match the performance of word-level language models. This approach has been adopted later by ByT5 (Xue et al., 2021a) where the text is represented by a sequence of utf-8 bytes. This allows us to represent all languages with a simple and small vocabulary of 256 symbols. English characters will be composed of single bytes while Arabic and Russian ones will consume 2-3 bytes per character. ByT5 comes in several capacities ranging from small to XXL, (0.3-13B) parameters respectively. Each of those were pre-trained on mC4 corpus which was crawled from web pages that cover 100 languages(Xue et al., 2021b). ByT5's authors showed that models processing language at byte level handle misspellings and noise gracefully and perform very well in languages with complex morphology.

| | Split | Examples ($\times 10^3$) | Words ($\times 10^6$) | Diacritics (%) |
|---|---|---|---|---|
| Tashkeela [21] | Train | 1750 | 75.7 | 78.0 |
| Tashkeela | Dev | 2.5 | 0.1 | 82.2 |
| Tashkeela | Test | 2.5 | 0.1 | 82.2 |
| CA [21] | Train | 1700 | 74.7 | 78.2 |
| MSA [21] | Train | 49 | 0.86 | 59.7 |
| Clean-50 [5] | Train | 50 | 2.1 | 83.1 |
| Extra [6] | Train | 533 | 22.6 | 82.2 |
| Clean-400 [8] | Train | 400 | 19.7 | 82.2 |

Table 2: Datasets variants extracted from Tashkeela corpus.

## 3 Diacritization Modeling

We consider the problem of predicting diacritics as a sequence-to-sequence task instead of a classification one. More specifically, our approach uses a text-to-text format: The input, fed to the model, consists of a sequence of utf-8 bytes where each unicode character is represented by 2-3 bytes subsequences. The model is asked to produce the same sequence interleaved with utf-8 bytes representing the diacritics. Note, that both input and output sequences are of variable length. For example, we are asking the model to produce the following sequence: `[217, 138, 217, 143, 217, 175, 217, 146, 217, 129, 217, 142, 217, 185, 217, 143]` which corresponds to يُدْفَعُ given the following input sequence: `[217, 138, 217, 175, 217, 129, 217, 185]` يدفع. This text based-input/output representation reduces the burden on the practitioner preparing datasets and integrating preprocessing, later on, in deployed models.

| Row | Model | Training Dataset | | Eval | Included Chars in Eval | | | | | DER | WER |
|---|---|---|---|---|---|---|---|---|---|---|---|
| | | Stage 1 | Stage 2 | Split | Numbers | Punct | Space | Last | Unlabeled | | |
| 1 | FineTashkeel / Small | CA | — | Dev | | | | ✓ | ✓ | 1.55 | 4.39 |
| 2 | | MSA | — | Dev | | | | ✓ | ✓ | 6.97 | 18.43 |
| 3 | | Clean-50 | — | Dev | | | | ✓ | ✓ | 1.70 | 4.73 |
| 4 | | Extra | — | Dev | | | | ✓ | ✓ | 1.35 | 3.74 |
| 5 | | Clean-400 | — | Dev | | | | ✓ | ✓ | 1.33 | 3.75 |
| 6 | | Tashkeela | — | Dev | | | | ✓ | ✓ | 1.38 | 4.02 |
| 7 | | Tashkeela | Clean-400 | Dev | | | | ✓ | ✓ | 1.16 | 3.35 |
| 8 | | Tashkeela | Clean-400 | Dev | | | | | ✓ | 0.98 | 1.97 |
| 9 | | Tashkeela | Clean-400 | Dev | | | | ✓ | | 1.31 | 3.09 |
| 10 | | Tashkeela | Clean-400 | Dev | | | | | | 1.13 | 1.91 |
| 11 | FineTashkeel / Base | Tashkeela | — | Dev | | | | ✓ | ✓ | 1.23 | 3.67 |
| 12 | | Tashkeela | Clean-400 | Test | | | | ✓ | ✓ | 1.00 | 2.92 |
| 13 | | Tashkeela | Clean-400 | Test | ✓ | ✓ | | ✓ | ✓ | 0.95 | 2.49 |
| 14 | | Tashkeela | Clean-400 | Test | ✓ | ✓ | ✓ | ✓ | ✓ | 0.74 | 2.49 |
| 15 | Barqawi (2017) | | | Test | | | | ✓ | ✓ | 3.73 | 11.19 |
| 16 | Fadel et al. (2019b) | | | Test | | | | ✓ | ✓ | 1.78 | 5.38 |
| 17 | Karim and Abandah (2021) | | | Test | ✓ | ✓ | | ✓ | ✓ | 1.97 | 5.13 |
| 18 | Madhfar and Qamar (2020) | | | Test | ✓ | ✓ | ✓ | ✓ | ✓ | 1.13 | 4.43 |

Table 3: Results on Tashkeela Validation and Test splits.

By describing the problem at a high level we are assigning the tasks of pre/post-processing and problem solving to the model to handle on its own, simplifying research and applications design.

## 4 Datasets

Tashkeela is a corpus of vocalized Arabic text that covers both Classical Arabic (CA) and Modern Arabic (MSA) with classical sources constituting the majority of the corpus. The text comes mainly from books that are crawled from the web (Zerrouki and Balla, 2017). While the dataset is quite instrumental in our modeling, there are several shortcomings that complicate our learning task such as: (a) Missing Diacritics (b) More than 2 diacritics on a single character (c) Inconsistent unicode representation (d) Annotated foreign language characters. Hence, each of the research efforts, that followed, devised a different filtering criteria to improve the training dataset quality. Table 2 shows several subsets of the original dataset that are generated by different rules of filtering.

## 5 Metrics

To measure our progress solving the task of diacritization we organize our metrics into two classes:

1. **Diacritic Error Rate** (`DER`): The percentage of unicode characters with which we predicted the wrong diacritic.
2. **Word Error Rate** (`WER`): The percentage of words with which have at least one character with an incorrect diacritic.

We define diacritics to be the unicode characters that are included in the unicode plane defined by the range (`0x064B-0x0652`). These characters will be removed from the input while any unicode character outside that range will stay as part of the input sequence. However, defining words represents a challenge since there is no word boundary identifier in the corpus. Moreover, Arabic language typically morphs several parts of speech into the same contiguous token. To simplify the computation of `WER`, we consider white spaces to be our word boundary despite its limitations.

## 6 Results & Discussion

**Finetuning Setup** We conducted initial experiments on the smallest dataset of Tashkeela (Clean-50) to find reasonable hyperparameters for our experiments by finetuning ByT5 small model on 8-TPUv2 cores for a maximum of 6000 steps with a batch size 256 per TPU core. We found that the optimal learning rate is $3 \times 10^{-3}$, sequence length to be 512 bytes. For more information check Appendices [A,B].

**Does data quality matter?** Table 3 (Rows: 1-6) shows that filtering the original dataset benefits the quality of the finetuning. For example, `DER` improved from 1.38% to 1.33% by only finetuning on **Clean-400** instead of the full dataset **Tashkeela**. However, aggressive filtering as being done in **Clean-50** reduces the size of the training dataset significantly, in this case, to 50K examples and therefore hurt the quality of the finetuned model increasing `DER` from ∼1.35% to 1.70%.

To take advantage of the diversity of the largest training datasets without being affected by the noisy examples that could be included, we devise the following curriculum learning schedule: (1) we finetune our pretrained model on the full dataset for 8K steps. (2) we, further, finetune for extra 4K steps on **Clean-400** This schedule will expose the model to diverse examples while narrowing definition of the task of diacritization to how it was demonstrated in the cleaner subset of training examples in **Clean-400**.

Table 3 (Rows: 6-7) shows that our curriculum is able to capture the best of both worlds. Large datasets that improve the coverage of the model domain and cleaner more targeted dataset that adhere better to the task definition. This sequential finetuning is able to reduce `DER` from 1.33 to 1.16.

**Does scale matter?** Table 3 (Rows: 6 vs 11) shows the results of finetuning **Base** ByT5 model which is slightly 2x larger than **Small**. We are able to reduce `DER` from 1.38% to 1.23% which is consistent with previously reported results that demonstrates improvements on downstream tasks as the pretrained model capacity increases (Hernandez et al., 2021; Wei et al., 2022).

**Do we benefit from self-supervised training?** Table 3 (Rows: 12-18) shows our model results in comparison to previously reported results on Tashkeela test dataset. Previous efforts evaluated their models with varying sets of characters. We evaluated our model consistently with each baseline (Rows: 12 $\leftrightarrow$ 16, 13 $\leftrightarrow$ 17, 14 $\leftrightarrow$ 18). Regardless of which evaluation methodology we have been using, we are able to reduce the error rate by at least 40% in `WER`.

**Are all diacritics hard?** Arabic grammar influences the last character diacritic of the word. Therefore, predicting the diacritic of the last character tends to be a harder problem since it depends on the relative position of the word within the sentence and its meaning. To test this hypothesis we evaluate our model excluding last character (Rows: 7 vs 8). We observe a drop in `DER` from 1.16% to 0.98% confirming that complexity of predicting the last characters. On the other hand, it seems that the model has an easier time identifying which characters should not be annotated with diacritics from the fact that `DER` increased from 1.16% to 1.31% (Rows: 7 vs 9).

## 7 Analysis

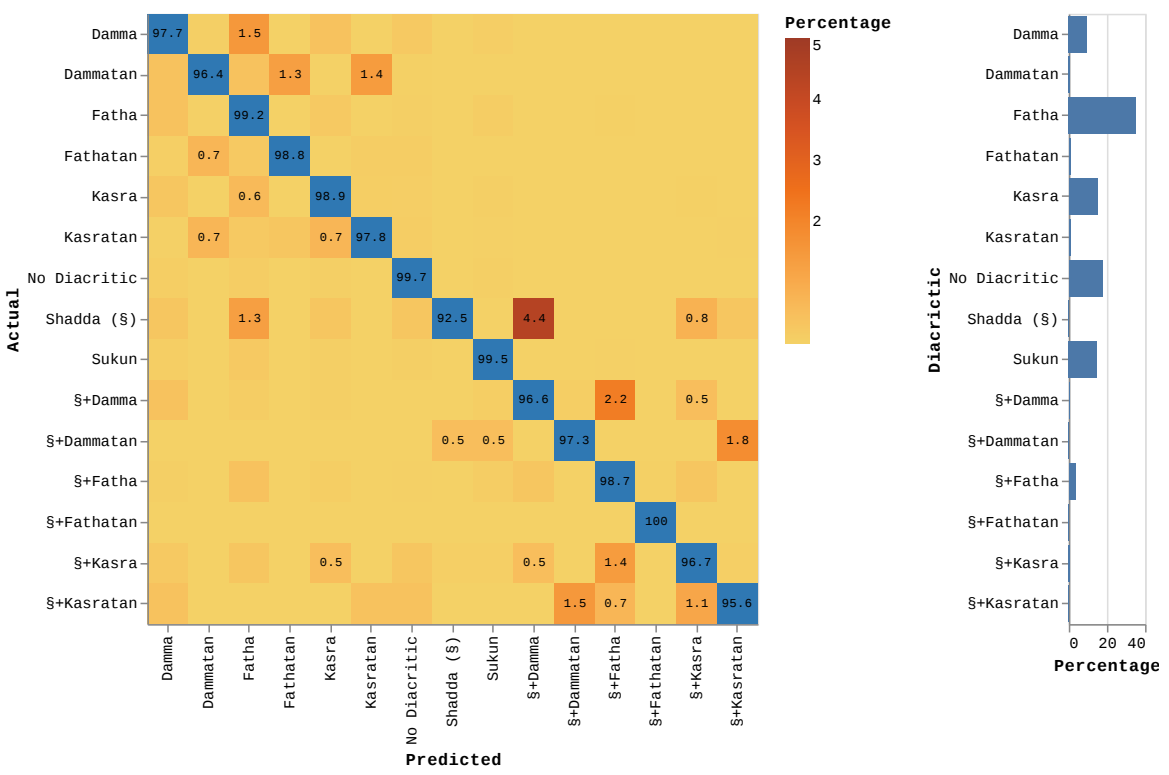

Figure 1: Confusion matrix of our model predictions on Tashkeela test dataset and the distribution of the diacritics.

To understand the categories of errors our model introduces, we calculated the confusion matrix in Figure 1. In the right side, we calculate the distribution of diacritics in our dataset. Single diacritics dominated the distribution while combined ones such as *Shadda + Fathatan* rarely appear. On the left, each cell represents the probability of predicting character at column (j) given the ground truth character (i) row. We notice that when the model is not quite certain it defaults to predicting Fatha. This could be explained by the fact it is the most common diacritic. Our model is performing very well (*No Diacritic* Accuracy=99.7) not predicting diacritics where they should not be (row) or missing predicting them when they are needed (column). Table 4 shows several errors done by our model.

| Target | Output | Issue |
|---|---|---|
| فِي فَرِيضَتَيْنِ بَدَلٌ | فِي فَرِيضَتَيْنِ بَدَلُ | Wrongs diacritic |
| حُمِلَ مِنْ مَجْلِس الْخِيَارِ | حُمِلَ مِنْ مَجْلِسِ الْخِيَارِ | Adding extra diacritic |
| وَبَنِيّ بَكْرٍ | وَبَنِي بَكْرٍ | Missing diacritic |

Table 4: Examples of our finetuned ByT5-Base model predictions on Tashkeela Dev.

## 8 Future Work & Conclusion

We have demonstrated that finetuning large pretrained multilingual models produces significant improvements in quality for automatic diacritization achieving new state-of-the-art results. We studied the benefits of scaling up the pretrained models and the impact of training dataset quality and size.

We realize that these pretrained models tend to be computationally expensive and not practical to be deployed in edge-compute applications. We are looking into distilling our best model into smaller models that are cheaper to run.

## A Learning Rate

In the following section, we determine the best learning rate to finetune ByT5-Small, with learning rates ranging from ($3 \times 10^{-4}$–$10^{-2}$). Figure A.1 shows that the learning rate of $3 \times 10^{-3}$ produces the best results for both word and character error rates.

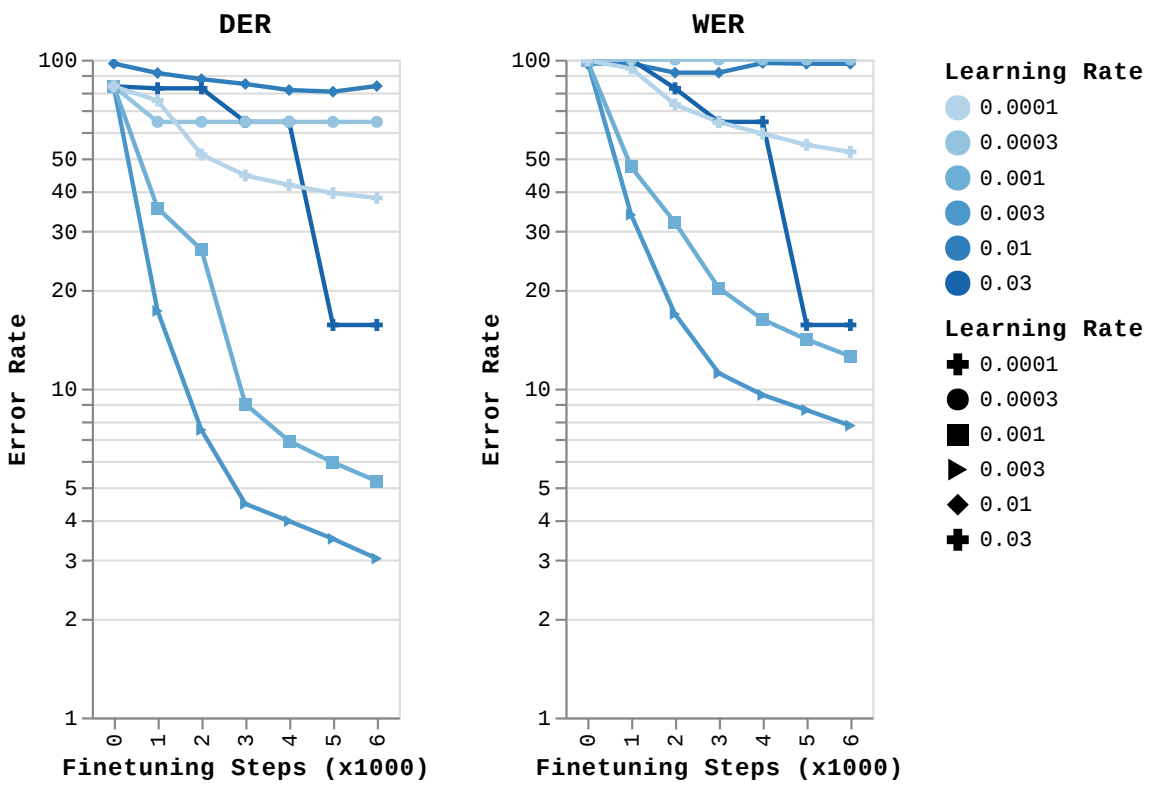

Figure A.1: Error rate of finetuning ByT5 Small for varying input lengths

## B Sequence Length

To make sure that we have a diverse sample of sequences of the training dataset, we investigate different maximum sequence lengths to finetune ByT5. ByT5 preprocessing logic truncates the sequences that are longer than a specific maximum length and packs more than one example in a single sequence if it is shorter than the maximum. This packing logic is meant to improve the utilization of computing budget during training and inference. For each example, we take the first N bytes in that sequence.

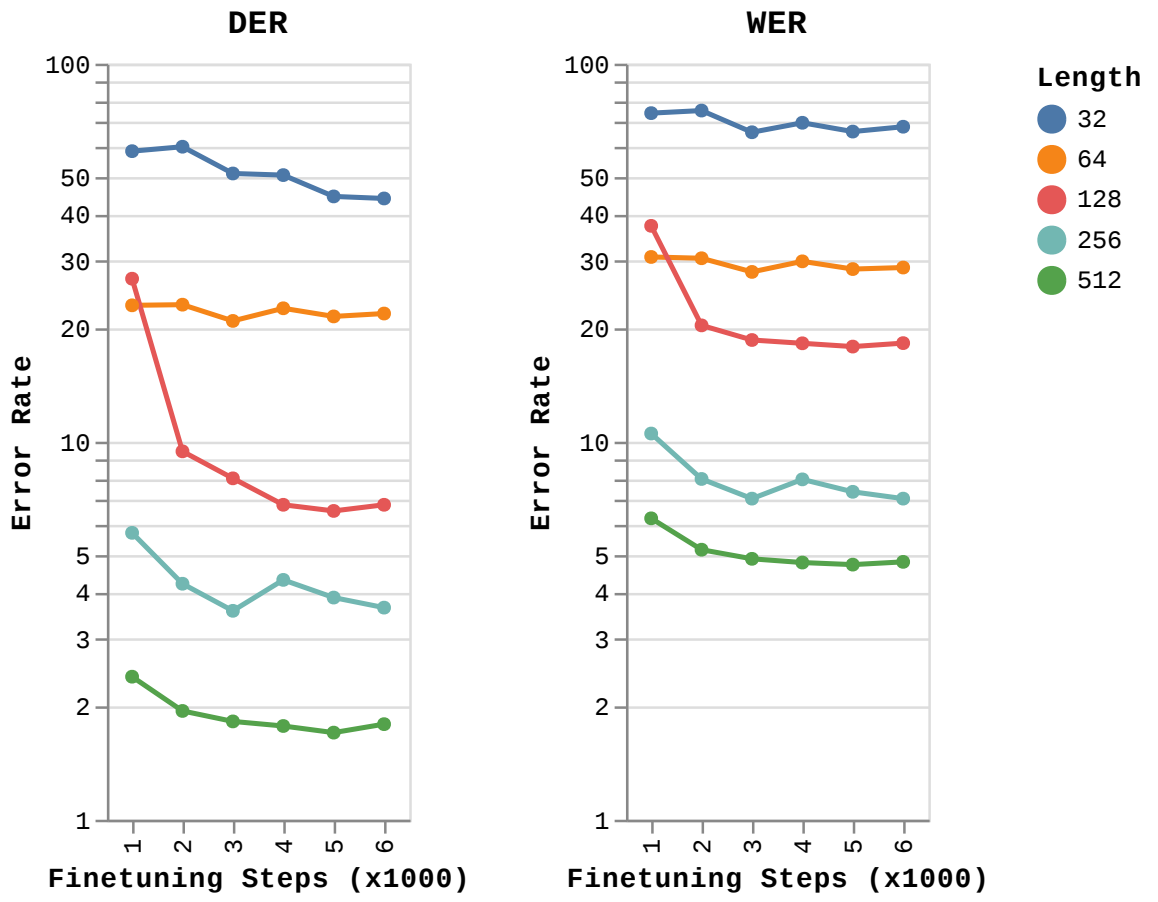

Figure B.1: Error rate of finetuning ByT5 small for varying input lengths

Figure B.1 shows that bigger sequences are necessary to avoid any negative effect of truncating sentences. Moreover, we show that with only 2000 steps of training sequences of length 512 bytes, we are able to achieve a DER less than 2%. For all of the experiments in the paper, we used a sequence length of 512 bytes.

## C Output Format

We investigate a different output format that replaces characters with a sentinel token. Our main concern is that we might be introducing errors in the character generation part which could have been easily avoidable if the task was modeled as a classification task. Therefore, we devised another output format for the problem, where the characters in the target sequence have been replaced by a sentinel token: "_". For this variation, we add a post-processing stage where we replace the sentinel token in the predicted sequence with the original example characters from the input sequence if both matched in length.

| Target Sequence | DER | WER |
|---|---|---|
| Characters & Diacritics | 1.7 | 4.7 |
| Sentinels & Diacritics | 4.4 | 19.2 |

Table C.1: Finetuned ByT5 model evaluated on Tashkeela Validation dataset.

While our current assumption that replacing all characters by a sentinel token should reduce the complexity of the problem since the model does not need to distinguish which character to output and just focus its compute budget on the diacritics positions, Table C.1 shows the opposite. We see that it is easier (lower error rates) for our finetuned model to predict a sequence that consists of both characters and diacritics. This might be explained by the fact that the model during pre-training has been tasked with generating natural sentences and our new synthesized sequences with the sentinel token are out of its domain and, therefore, harder to learn.

## D Effect of Frequency

Figure D.1 shows a power law relationship between the frequency of a diacritic appearance and its error rate. The more frequently the model has been exposed to a specific diacritic,

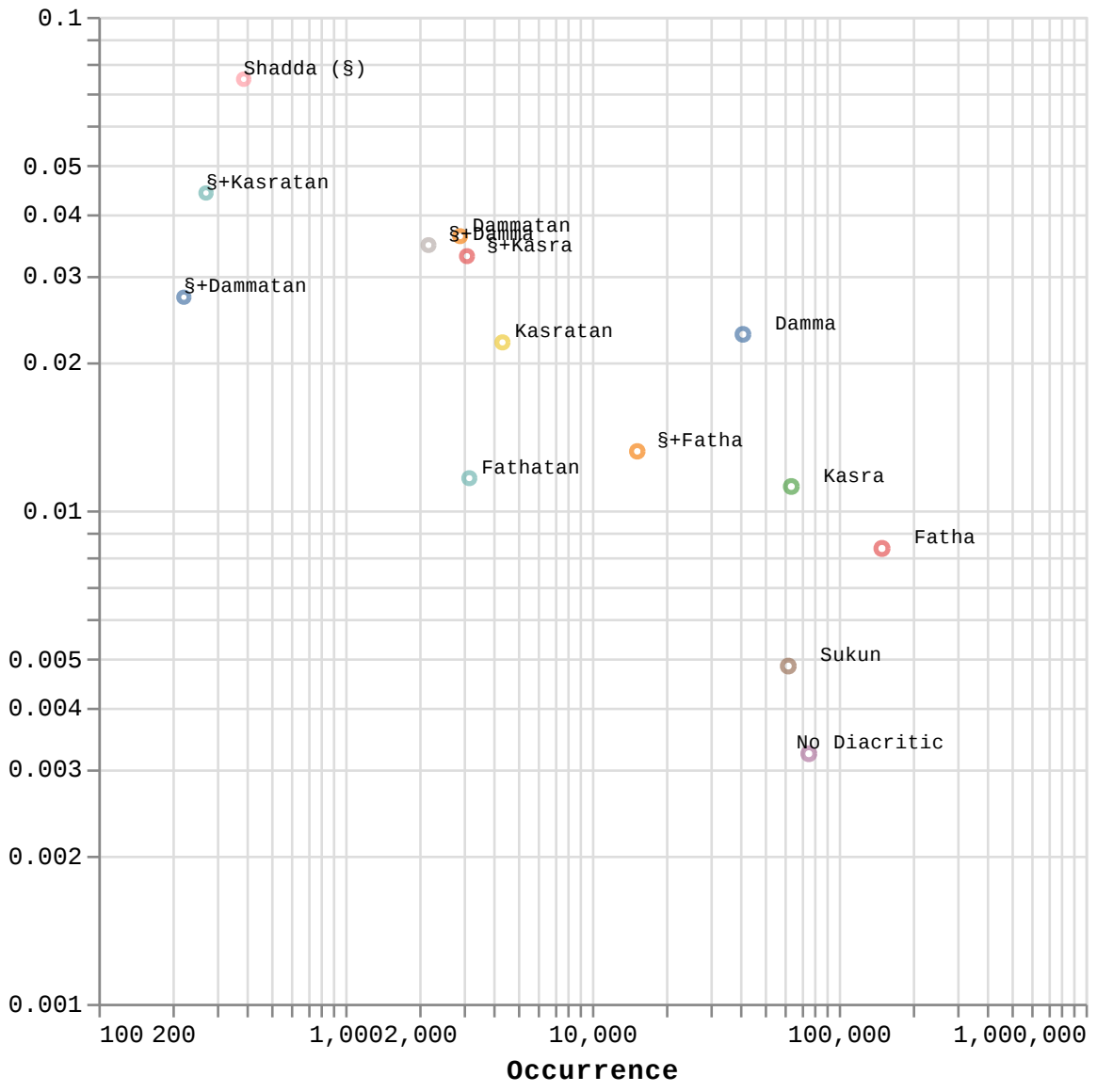

Figure D.1: Relation between frequency of the diacrtic occurrence and its error rate.

the more accurate the predictions. This might hint onto future directions where we oversample the diacritics that appear naturally less frequently such as combined diacritics that include *Shadda (§)*.